\documentclass{article}
\usepackage[preprint, nonatbib]{neurips_2024}

\usepackage[utf8]{inputenc} 
\usepackage[
T1]{fontenc}    
\usepackage{hyperref}       
\usepackage{url}            
\usepackage{booktabs}       
\usepackage{amsfonts}       
\usepackage{nicefrac}       
\usepackage{microtype}      
\usepackage{xcolor}         

\DeclareMathAlphabet\mathbfcal{OMS}{cmsy}{b}{n}

\def\0{{\bf 0}}
\def\1{{\bf 1}}

\def\eg{\emph{e.g.}} 
\def\ie{\emph{i.e.}} 
 
 \def\vs{\emph{vs.}}
\def\wrt{{w.r.t.~}}

\usepackage{amsmath}

\usepackage{subfigure}
\usepackage{graphicx}
\DeclareMathAlphabet\mathbfcal{OMS}{cmsy}{b}{n}
\usepackage{algorithm}
\usepackage{algorithmic}
\usepackage{amsmath}
\usepackage{multirow}
\usepackage{bbding}
\usepackage{caption}
\newcommand\tabcaption{\def\@captype{table}\caption}
\newcommand\figcaption{\def\@captype{figure}\caption}

\newcommand{\topline}{\toprule [0.1em]}
\newcommand{\midline}{\midrule [0.05em]}
\newcommand{\bottomline}{\bottomrule [0.1em]}

\def\icml{\textcolor{black}}	
\def\task{few-shot audio-visual acoustics modeling}
\def\Task{Few-shot audio-visual acoustics modeling}

\def\shortname{MAGIC}

\def\originmap{observation semantic map}

\def\HOriginMap{Observation semantic map} 
\def\originmapshortname{OSM}
\def\anticipatedmap{scene semantic map}
\def\Hanticipatedmap{Scene semantic map}

\def\anticipatedmapshortname{SSM}

\def\baseline{Few-ShotRIR}

\usepackage{todonotes}

\usepackage{caption} 

\hypersetup{
 colorlinks=true,
 linkcolor=red,
 citecolor=cyan,
 filecolor=magenta, 
 }

\title{\raisebox{-0.238\height}{\includegraphics[width=5.8em]{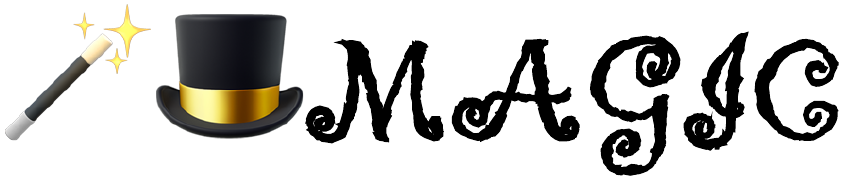}}: Map-Guided Few-Shot Audio-Visual \\ Acoustics Modeling}

\author{
    Diwei Huang\textsuperscript{\rm 1}\thanks{Equal contribution.}~~
    Kunyang Lin\textsuperscript{\rm 1}\footnotemark[1]~~
    Peihao Chen\textsuperscript{\rm 1}~~
    Qing Du\textsuperscript{\rm 1}~~  
    Mingkui Tan\textsuperscript{\rm 1}\thanks{Corresponding author.}~~  \\
    \textsuperscript{\rm 1}South China University of Technology, \\
    \{sediweihuang, imkunyanglin\}@gmail.com, mingkuitan@scut.edu.cn
}

\begin{document}

\maketitle
\vspace{-7mm}

\begin{abstract}
\Task{} seeks to synthesize the room impulse response in arbitrary locations with few-shot observations.
To sufficiently exploit the provided few-shot data for accurate acoustic modeling, we present a \textit{map-guided} framework by constructing acoustic-related visual semantic feature maps of the scenes. Visual features preserve semantic details related to sound and maps provide explicit structural regularities of sound propagation, which are valuable for modeling environment acoustics. We thus extract pixel-wise semantic features derived from observations and project them into a top-down map, namely the \textbf{\originmap{}}. This map contains the relative positional information among points and the semantic feature information associated with each point. Yet, limited information extracted by few-shot observations on the map is not sufficient for understanding and modeling the whole scene. We address the challenge by generating a \textbf{\anticipatedmap{}} via diffusing features and anticipating the \originmap{}. The \anticipatedmap{} then interacts with echo encoding by a transformer-based encoder-decoder to predict RIR for arbitrary speaker-listener query pairs. Extensive experiments on Matterport3D and Replica dataset verify the efficacy of our framework.
\end{abstract}

\section{Introduction}
\Task{} aims to predict the room impulse response (RIR) capturing the process of sound propagation at arbitrary query locations by exploiting given support few-shot RGB-D observations and echo information from a 3D scene~\cite{majumder2022few}. This task is a kind of acoustic synthesis~\cite{chen2023novel, chen2023everywhere, lee2022hierspeech, kong2020hifi} that focuses on spatial learning. Its core objective is to comprehend the entire acoustic space and subsequently forecast the RIR at any specified location. 

As a foundational capability in acoustic understanding, \task{} provides powerful spatial awareness for downstream tasks, such as audio-visual navigation~\cite{chen2020soundspaces, younes2023catch, wang2023learning, chen2020learning}. Moreover, the application of \task{} has been extended to diverse real-world applications, including sound simulation in virtual reality (VR)~\cite{postma2015creation, remaggi2019reproducing, okazawa2023auditory, yilmazer2023virtual}, perceptual enhancement in augmented reality (AR)~\cite{gari2019flexible, bona2022automatic, liang2023reconstructing}, and optimization of room acoustics in architectural design~\cite{elkhateeb2012acoustical, kummritz2019acoustic, yu2020room}. Driven by such practical needs, \task{} has garnered significant attention. 

\begin{figure*}[!t]
    \centering
	\includegraphics[width=1.0\linewidth]{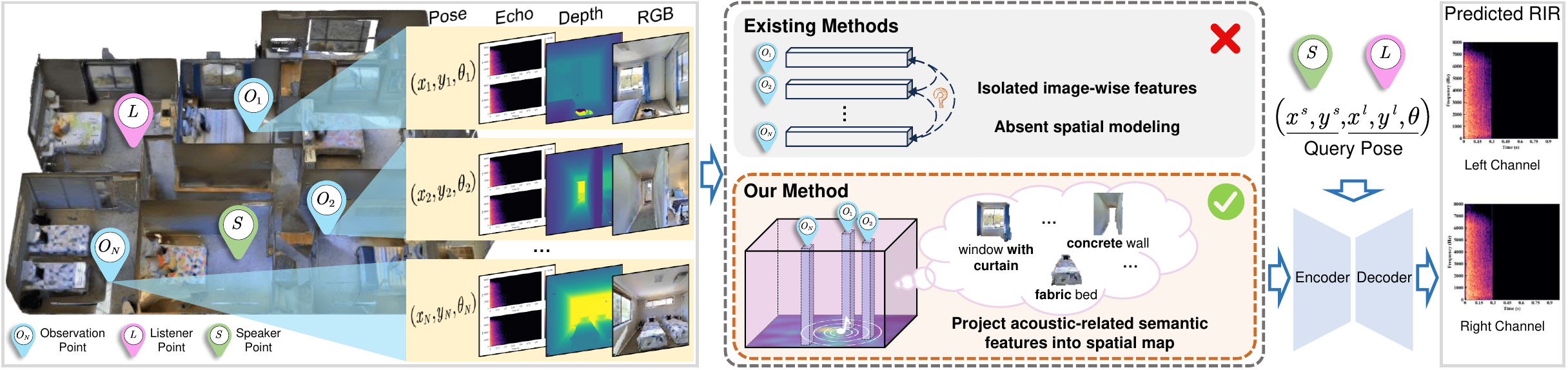}
	\caption{Illustration of few-shot audio-visual acoustics modeling and the distinguishment between existing methods and our method. Existing methods directly extract visual features from the image-wise encoder and feed them to the RIR prediction module. In contrast, our method builds a feature map that captures pixel-wise semantic context related to sound production to help acoustic modeling.}
	\vspace{-4mm}
    \label{fig:comparison}
\end{figure*}

Despite these advances, it is difficult to sufficiently understand and exploit the provided few-shot data thus limiting the performance of acoustic learning. Previous works have explored promising attempts. They mainly perceive the environment from raw RGB-D images and predict target RIR in an end-to-end manner. Yet, these methods typically overlook the spatial and semantic relationship among the support observations. They either implicitly model the local features by Nerf~\cite{luo2022learning} without utilizing the full visual information, especially semantic information, or weaken the pose information with a sinusoidal positional encoding without considering the connection between each location~\cite{majumder2022few}. 

We notice that environmental materials intricately influence the process of sound transmission. As sound is emitted, it interacts with objects in the environment—absorbed, scattered, and transmitted—before reaching the listener~\cite{chen2022soundspaces}. Objects with different shapes and materials have varying acoustic properties, resulting in distinct auditory experiences. Fortunately, fine-grained details and semantic information that can infer the shapes and materials of the objects have been demonstrated to be contained in the visual features from the semantic segmentation network U-Net~\cite{chen2022weakly}. Motivated by this observation, we resort to pixel-wise visual object features to help the model perceive and understand the environments from the acoustic perspective.

In addition to materials, the propagation of sound is influenced by both the emitted location and the received location. Therefore, it is important to learn the model to exploit structural and semantic regularities of real-world environments. A very intuitive idea is to map all known information from different locations on a comprehensive map representing the whole scene. As shown in Figure~\ref{fig:comparison}, compared to the existing method~\cite{majumder2022few} that directly inputs image-wise features (\eg~ResNet~\cite{he2016deep}), feature maps constructed by pixel-wise semantic features (\eg~U-Net~\cite{ronneberger2015u}) provide information related to the sound production as well as specific positional relationships among the features. 

Yet, the observations we use to construct a map are limited in the few-shot learning. This results in restricting the performance gains from the map to the provided few-shot viewpoints. In practice, there exists a correlation between seen and unseen regions; for instance, observing a corner of an object suggests the presence of other parts with similar acoustic properties nearby~\cite{ramakrishnan2020occupancy}. Instead of passively awaiting additional observations from external sources, it is more advantageous to empower the model to deduce scene characteristics in unseen regions based on feature context. This actively anticipating approach enables the model to go beyond the visible cues, facilitating more precise predictions of RIR at unknown locations.

In light of these findings, we propose to construct the semantic visual feature map to guide the few-shot audio-visual acoustics modeling. We name our method as \underline{ma}p-\underline{g}uided few-shot audio-v\underline{i}sual acousti\underline{c}s modeling~(\textbf{MAGIC}). The key ingredient of our approach is to construct the visual feature map in novel environments to perceive sound propagation representation within the scene. To accomplish this, we extract pixel-wise semantic features of objects and project them by depth projection into a top-down map referred to as the \textbf{\originmap{}}. This map contains the relative positional information among points and the semantic feature information associated with each point. To deal with the sparsity of features in the \originmap{} constructed from few-shot images, we introduce the feature anticipation module. This module analyzes the characteristics of the few-shot viewpoints and anticipates semantic features beyond the seen regions for the entire scene, generating the \textbf{\anticipatedmap{}}. The incorporation of this module enhances the model's perception of the overall scene, particularly in unseen regions. Finally, we segment the feature map into multiple patches and fuse these patches with audio features to predict the RIR.

Our approach is rigorously validated through extensive experiments on 83 Matterport3D~\cite{chang2017matterport3d} scenes and 18 Replica scenes~\cite{straub2019replica} from SoundSpaces~\cite{chen2020soundspaces}. Experimental results show that our model outperforms the state-of-the-art model. Promising results demonstrate that \shortname{} is able to learn spatial and semantic properties of the environment and thus help few-shot acoustic learning.

To sum up, our main contributions are as follows: 1) We structurally represent the information of different locations to facilitate the model to understand the scene structure, layout, semantics, etc., to determine the sound propagation characteristics accurately. 2) We introduce the feature anticipation module to strengthen the connection between features and predict features at unseen points with given observations, thus improving the accuracy of RIR prediction for any point. 3) Experimental results on Matterport3D and Replica demonstrate the superior performance of our method compared to state-of-the-art methods. \shortname{} achieves better results with less training data \textit{\wrt} both the number of scenes and the given observations in each scene.

\section{Related work}

\paragraph{Acoustic synthesis.}
Acoustic synthesis can be categorized into different classes, such as speech acoustic synthesis~\cite{kong2020hifi, lee2022hierspeech, chen2023novel, gan2020foley, wu2022generating} and RIR acoustic synthesis~\cite{luo2022learning, majumder2022few, liang2023neural, ratnarajah2023av}.  In this paper, we focus on RIR acoustic synthesis that models the acoustic characteristics of environments instead of speech perception and production (\ie~speech acoustic synthesis). Early works~\cite{murphy2007acoustic, holters2009impulse} rely on physical formulas to derive sound field distributions. Recently, Fast-RIR~\cite{ratnarajah2022fast} takes acoustic parameters RT60 as inputs and generates both specular and diffuse reflections for a given acoustic environment. Similarly, several works~\cite{eaton2016estimation, gamper2018blind, ratnarajah2020ir} use acoustic parameters to optimize RIR predictions. NAF~\cite{luo2022learning} exploits Nerf~\cite{mildenhall2021nerf} to model the whole scene and predict the RIR using a few viewpoints but requires training individual models for each scene. To address this, \baseline{}~\cite{majumder2022few} used the attention mechanism of transformer~\cite{vaswani2017attention} to learn the potential correlation between visual features and audio features. However, directly extracting features and inputting them into the transformer ignores explicit connections between information (\eg, relative relationships between viewpoints, correspondence between images and poses) and does not fully utilize the semantic information. In this paper, we extract acoustically relevant semantic features from the observations and project them to a top-down map to help generalize the model to unseen locations.

\paragraph{Map construction.} 
Map construction is a common way to store scene information and perceive the environment in various tasks, such as visual navigation ~\cite{ramakrishnan2020occupancy, chaplot2019learning, chen2022learning}, visual language navigation~\cite{anderson2018vision, chen2022weakly, lily}, and audio-visual navigation~\cite{chen2020soundspaces, younes2023catch, wang2023learning, chen2020learning}. In visual navigation, ANS \cite{chaplot2019learning} and OccAnt~\cite{ramakrishnan2020occupancy} utilize obstacle maps indicating whether a point is occupied to improve the agent's exploration efficiency. In visual language navigation, WS-MGMap~\cite{chen2022weakly} constructs a multi-granularity map to represent fine-grained details. In audio-visual navigation, AV-WaN~\cite{chen2020learning} proposes an audio intensity map to store the acoustic memory. In contrast to these works, we construct an acoustic-related spatial semantic map for \task{} to improve the model's understanding of the scene.

\section{Problem definition}

Given several egocentric visual-echo observation pairs and the corresponding positions where the echoes are heard, \task{} aims to query the RIR of arbitrary speaker-listener poses through acoustics modeling of the scenes. 

We represent $\mathbfcal{O}$ as the egocentric audio-visual observations randomly sampled from a 3D environment, which can be regarded as the support set in few-shot learning. The poses to be queried are represented by $\mathbfcal{Q}$, which can be viewed as the query set in few-shot learning. 
We let $\mathbfcal{R}$ denote the RIR that is corresponding to $\mathbfcal{Q}$. 

In this paper, our objective is to learn a function $f$ to predict the RIR for the arbitrary query $\mathbfcal{Q}$ given the egocentric audio-visual observations $\mathbfcal{O}$:

\begin{equation}
\mathbfcal{R} = f(\mathbfcal{Q};\mathbfcal{O}).
\end{equation}

Specifically, $\mathbfcal{O}$ consists of visual observations $\mathbfcal{V}$, echo observations $\mathbfcal{A}$, and pose observations $\mathbfcal{P}$. The total number of provided positions is $N$, which means that $\mathbfcal{O} = \{O_i\}^N, \mathbfcal{V} = \{V_i\}^N, \mathbfcal{A} = \{A_i\}^N, \mathbfcal{P} = \{P_i\}^N$. $V_i$ and $A_i$ represent the egocentric RGB-D view and the binaural echo response in $i^\text{th}$ position, respectively. A pose $P_i$ includes the coordinates of the speaker $(x^s_i, y^s_i)$ and listener $(x^l_i, y^l_i)$, along with the orientation of the listener denoted as $\theta_i$. Formally, $P_i = (x^s_i, y^s_i, x^l_i, y^l_i, \theta_i)$ with the constraint $x^s_i = x^l_i$ and $y^s_i = y^l_i$. Note that while the speaker is omnidirectional, the listener receives input in four orientations, \ie~$\theta_i \in \{{0^{\circ}, 90^{\circ}, 180^{\circ}, 270^{\circ}}$\}.

The total number of query positions is $N'$, such that $\mathbfcal{Q} = \{Q_j\}^{N'}, \mathbfcal{R} = \{R_j\}^{N'}$. During the querying process, $Q_j = (x^s_j, y^s_j, x^l_j, y^l_j, \theta_j)$ represents the query coordinates, and the corresponding RIR function is predicted as $R_j$. It is essential to note that most of the query positions do not coincide with the positions in the support set. This increases the challenge of the task.

\section{Proposed methods}

\begin{figure*}[t]
    \centering
	\includegraphics[width=1.0\linewidth]{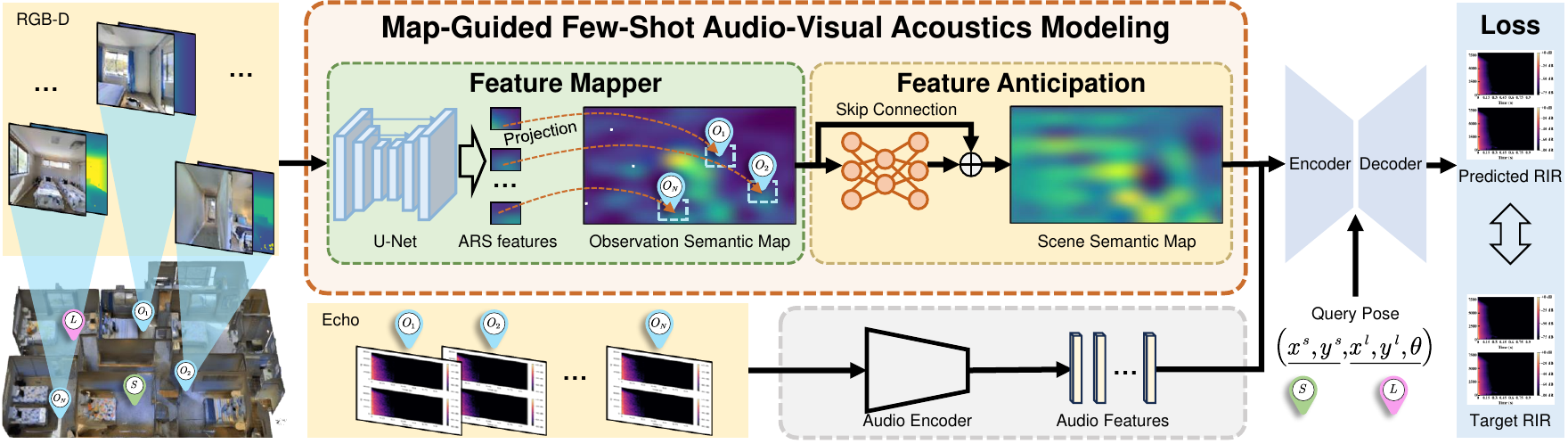}
	\caption{General scheme of \shortname{}. \shortname{} leverages U-Net pre-trained by semantic segmentation to extract the acoustic-related semantic (ARS) features and project the pixel-wise features to the \originmap{}. Then, the \originmap{} is fed into feature anticipation module to anticipate the unseen area features. The resulting \anticipatedmap{} interacts with echo encoding by a transformer-based encoder-decoder to predict RIR for arbitrary speaker-listener query pairs. We train the model minimizing the loss between the predicted RIR and target RIR.
	}

	\vspace{-3.8mm}
    \label{fig:overview}
\end{figure*}

We aim to learn an RIR prediction model for arbitrary queries by exploiting visual feature maps to model the environment acoustics. As shown in Figure~\ref{fig:overview}, the visual branch and the audio branch that takes echo as input constitute the entire few-shot audio-visual acoustics modeling pipeline. 

In the visual branch, we propose two key maps: 1) \textit{\originmap{}}, and 2) \textit{\anticipatedmap{}}. The \originmap{} captures the spatial and semantic acoustic characteristics of observed regions by extracting pertinent information from the support observations (Sec.\ref{sec:\originmap{}}). The \anticipatedmap{} is derived through the anticipation of the \originmap{}, aiming to predict the acoustic properties of query unseen regions while preserving the acoustic attributes of seen areas (Sec.\ref{sec:\anticipatedmap{}}).

The audio branch incorporates the pose information into the audio observations to obtain audio features. Finally, the outputs of the visual branch and audio branch concatenate together and are fed into the encoder-decoder to predict the final RIR. An overview of the training process is presented in Algorithm~\ref{alg:training_algorithm} in supplementary materials.

We next introduce how to generate these two types of map representation and utilize them in few-shot learning of environment acoustic.

\subsection{\HOriginMap{} construction}
\label{sec:\originmap{}}
Observations sampled from an environment store both intrinsic spatial connections and semantic features closely associated with the sound propagation process. Based on this observation, we propose leveraging fine-grained semantic features on objects to build an \originmap{} for perceiving the environment. Given visual observations $\mathbfcal{V}$ and pose observations $\mathbfcal{P}$, as shown in Figure~\ref{fig:overview}, we extract acoustic-related semantic features using pre-trained U-Net~\cite{ronneberger2015u} and project the final-layer features to construct the top-down map, generating the \originmap{} $M_{\mathrm{\originmapshortname{}}}$. 

\textbf{Feature extraction.} Objects with different shapes and materials have varying acoustic properties, naturally resulting in distinct auditory experiences. Previous research ~\cite{islam2020shape, zhou2018interpreting, liu2020dynamic} has emphasized that features at different levels of the model encompass various fine-grained details of the objects. Specifically, low-level features capture shape and texture, while high-level features encompass various parts of the object. Based on these insights, we exploit a pre-trained U-Net ~\cite{ronneberger2015u} semantic segmentation model as the visual feature extraction encoder $f_V$. The raw RGB images of visual observations $\mathbfcal{V}$ captured at location $\mathbfcal{P}$ are fed into the encoder $f_V$. The output consists of potential features from the final layer of the U-Net network. Leveraging the jump-connection mechanism of U-Net, this final layer fuses both low-level and high-level features, providing a comprehensive representation of environmental details. More details are provided in the supplementary materials. Consequently, we attain pixel-wise acoustic-related semantic features $V$:
\begin{equation}
    V = f_V(\mathbfcal{V}),
\end{equation}
where $V \in \mathbb{R}^{N \times h \times w \times c_f}$, $h$ and $w$ denote the height and width of images, respectively. $c_f$ signifies the number of channels corresponding to the acoustic-related semantic features.

\textbf{Map construction.} Visual features $V$ offer a diverse and detailed acoustic characterization of the environment. However, directly concatenating them into a scene memory may overlook the spatial connections between these visual features and is limited to the local horizon. In addressing the aforementioned issues, we project visual features from various locations onto a unified egocentric top-down map~\cite{cartillier2021semantic}, utilizing depth mapping operation $m(\cdot)$ based on the depth map and pose information $\mathbfcal{P}$. \icml{More details can be found in supplementary materials.} We represent the resulting projected map as the \originmap{}, denoted as $M_{\mathrm{\originmapshortname{}}}$:
\begin{equation}
    \label{eqa:mapping}
    M_{\mathrm{\originmapshortname{}}} = m(V, D, \mathbfcal{P}),
\end{equation}
where $M_{\mathrm{\originmapshortname{}}} \in \mathbb{R}^{m \times m \times c_f}$, $m \times m$ represents the map size, $D$ represents the depth images. Such constructed feature maps greatly improve the presentation of information and foster the utilization of available few-shot representations.

\subsection{\Hanticipatedmap{} construction}
\label{sec:\anticipatedmap{}}
\textbf{Feature anticipation.} The \originmap{} $M_{\mathrm{\originmapshortname{}}}$ is constructed using seen observations and lacks spatial perception of unseen regions. Our solution is to infer features for unseen regions based on the context of seen regions. As illustrated in Figure~\ref{fig:overview}, we propose a feature anticipation module, denoted as $f_{FA}$ to anticipate the features of unseen regions. To ensure consistency in all dimensions between pre- and post-prediction maps, we employ the U-Net-like network as the structure of the feature anticipation module. The \originmap{} generated in Section~\ref{sec:\originmap{}} is first to be downsampled for compressing the image size and expanding the dimension of channels, followed by a restoration of dimensionality through upsampling. Note that the output of the feature anticipation module will efficiently yield broader information about the entire scene. We represent it as the \anticipatedmap{}, denoted as $M_{\mathrm{\anticipatedmapshortname{}}}$:
\begin{equation}
    M_{\mathrm{\anticipatedmapshortname{}}} = f_{FA}(M_{\mathrm{\originmapshortname{}}}),
\end{equation}
where $M_{\mathrm{\anticipatedmapshortname{}}} \in \mathbb{R}^{m \times m \times c_f}$. Specifically, $f_{FA}(\cdot)$ represents the feature anticipation module. We employ skip connection followed ~\cite{he2016deep} to sum the input $M_{\mathrm{\originmapshortname{}}}$ and output of the learnable U-Net to generate the final output $M_{\mathrm{\anticipatedmapshortname{}}}$. Note that the skip connection operation has been viewed as a common practice~\cite{chen2020generating} to expand features while retaining original input features effectively and to reduce the learning difficulty. \icml{More details are provided in the supplementary materials. }

\textbf{Map embedding.} As shown in Figure~\ref{fig:overview}, we finally encode the \anticipatedmap{} to embedding features and feed it with the audio features obtained from the audio branch into the transformer-based encoder-decoder to predict
RIR for arbitrary speaker-listener query pairs. Specifically, following ViT ~\cite{dosovitskiy2020vit}, we split the map into multiple patches via convolution as embedding features. The features are combined with echo features in the subsequent stages to synthesize RIR.

\subsection{Learning objectives and optimization}
Following standard settings~\cite{majumder2022few}, we adopt STFT loss and energy decay matching loss as the learning objectives of \shortname. 

\textbf{STFT loss.} Given RIR prediction $P$ and the target $R$, $L_{\mathrm{STFT}}$ is $L_1$ loss in time-frequency domain:
\begin{equation}
    L_{\mathrm{STFT}} = \frac{1}{2 \times F \times T} \sum_{i=1}^{2 \times F \times T}{\lVert {P_i - R_i} \rVert _1},
\end{equation}
where $2$ is the number of channel, $F$ is the number of frequency bins, $T$ is the number of overlapping time windows. 

\textbf{Energy decay matching loss.} Instead of calculating the average prediction error, energy decay matching loss $L_{\mathrm{EDM}}$ focuses on the reverberation quality:
\begin{equation}
    L_{\mathrm{EDM}} = \frac{1}{2 \times T} \sum_{i=1}^{2 \times T}{\lVert D_\epsilon(P_i) - D_\epsilon(R_i) \rVert _1},
\end{equation}
where $D_\epsilon$ represents the function to calculate the decay curve by Schroeder’s backward integration algorithm. Note that the calculation does not take into account the case of those temporal positions where the target energy decay $D_\epsilon(R_i)$ is zero.

\textbf{Final loss.} The final loss $L$ is computed by weighting the two loss:
\begin{equation}
    L = L_{\mathrm{STFT}} + \lambda L_{\mathrm{EDM}},
    \label{eq:loss}
\end{equation}
where $\lambda$ is a balancing weight for $L_{\mathrm{EDM}}$ and set to $0.01$ by default.

\section{Experiment}
In this section, we first evaluate the superiority of \shortname{} on the \task{} task. We then verify the effectiveness of each module in \shortname{}. We also investigate the effect of different hyperparameters. Last, we provide qualitative comparison results.

\subsection{Experimental setup} 
\label{sec:es}

\textbf{Simulator environment.} We evaluate our method on SoundSpaces ~\cite{chen2020soundspaces} dataset. This dataset is built on the Habitat simulator ~\cite{savva2019habitat} and can deliver perceptually realistic 3D audio-visual scene simulations. SoundSpaces provides numerous location points for each scene and is convenient for using any two points as speaker and listener positions for sound rendering. Notably, the speaker is omnidirectional, while the listener is stereo and can be oriented in four distinct directions $\{0^{\circ}, 90^{\circ}, 180^{\circ}, 270^{\circ}\}$.

\textbf{Dataset setup.}  We evaluate our \shortname{} on 83 Matterport3D scenes ~\cite{chang2017matterport3d} and 18 Replica scenes~\cite{straub2019replica}. Following the \baseline{} ~\cite{majumder2022few}, the scenes are divided into \textit{seen} split and \textit{unseen} split, containing 56 scenes and 27 scenes respectively in Matterport3D. Considering the time efficiency, we construct a mini dataset derived from the Matterport3D dataset to ablate our proposed method, randomly reducing \textit{seen} split to 12 but remaining \textit{unseen} split unchanged. Replica contains 9 seen scenes and 9 unseen scenes. Replica has a higher spatial resolution (0.5m) than Matterport 3D (1m). More details are provided in the supplementary materials.

\textbf{Evaluation metrics.} Following standard setting ~\cite{majumder2022few}, we evaluate the quality of the generated RIR on four acoustic metrics, namely Short Time Fourier Transform Error (\textbf{STFT}), RT60 Error (\textbf{RTE}), DRR Error (\textbf{DRRE}) and Mean Opinion Score Error (\textbf{MOSE}). STFT is the error in calculating the L1 distance between the spectrogram of the predicted RIRs and the target RIRs. RTE is the error between the RT60 metric of predicted RIRs and the target RIRs. DRRE is the error in the estimated energy ratio between the direct and reverberant sounds of the predicted RIRs and target RIRs. MOSE measures the difference in the perceived quality of predicted RIRs and the target RIRs when convolved with human speech with the help of a deep learning network.

\textbf{Implementation details.} Our approach is implemented by using the Pytorch framework ~\cite{paszke2019pytorch}. We used 8 NVIDIA GeForce RTX 3090 GPUs (each with 24GB of memory) for training with a batch size of 24. The provided observations context size $N$ for both training and testing is set to 20.The number of queries $N'$ is 60 for training and 50 for testing. The map size $m$ of both the \originmap{} and the \anticipatedmap{} we constructed is 64 and the resolution is 1. More details are provided in the supplementary materials.

\subsection{Main results}

\textbf{Existing methods and baselines.}
We compare our mehtod with the state-of-the-art methods, \ie~Fast-RIR++~\cite{ratnarajah2022fast} and \baseline{} ~\cite{majumder2022few}. Following \baseline{}, we also compare other three baselines, \ie~Nearest Neighbor, Linear Interpolation, and AnalyticalRIR++:

$\bullet$ \textbf{\baseline{}} directly encodes the provided visual observations by ResNet and fuses the visual feature with pose embedding and modality embedding by linear layers. The other parts of the model architecture remain the same as our method.

$\bullet$ \textbf{Nearest Neighbor} computes the distance between the listeners of the query viewpoints and the support echo positions and outputs the echo RIRs that are closest to the listeners.

$\bullet$ \textbf{Linear Interpolation} finds out the top four closest support viewpoints to the listeners of query viewpoints and outputs the linear interpolation of their RIRs of echoes.

$\bullet$ \textbf{AnalyticalRIR++} is modified from \baseline{} and replaces the transposed convolutions with fully-connected layers for RT60 and DRR prediction. It analytically shapes an exponentially decaying white noise~\cite{lebart2001new} based on the predicted RT60 and DRR to estimate the target RIR.

$\bullet$ \textbf{Fast-RIR++} is based on Fast-RIR~\cite{ratnarajah2022fast} that trains a GAN\cite{goodfellow2014generative} to synthesize RIRs and require the environment and acoustic attributes to be known as a prior. Fast-RIR++ is constructed from AnalyticalRIR++, using panoramic depth images at the query speaker and listener to infer the scene size and training the model by augmenting the objective with the $L_{\mathrm{EDM}}$ loss.

\begin{table}[t!]
\centering
\caption{RIR prediction results on Matterport3D. All metrics use base $10^{-2}$ and lower is better.}
\label{tab:results}
\resizebox{0.8\linewidth}{!}{\begin{tabular}{lcccc|cccc}
    \topline
    \multirow{2}[2]{*}{Model} & \multicolumn{4}{c|}{Seen} & \multicolumn{4}{c}{Unseen} \\
    \multicolumn{1}{l}{} & STFT↓ & \multicolumn{1}{c}{RTE↓} & \multicolumn{1}{c}{DRRE↓} & \multicolumn{1}{c|}{MOSE↓} & \multicolumn{1}{c}{STFT↓} & \multicolumn{1}{c}{RTE↓} & \multicolumn{1}{c}{DRRE↓} & \multicolumn{1}{c}{MOSE↓} \\
    \midline
    Nearest Neighbor & 4.65  & 1.15  & 385   & 24.4  & 4.87  & 1.26  & 391   & 28.0 \\
    Linear Interpolation & 4.44  & 1.22  & 393   & 24.3  & 4.67  & 1.32  & 403   & 27.2 \\
    AnalyticalRIR++ & 2.94  & 0.98  & 463   & 28.1  & 3.02  & 1.19  & 467   & 29.4 \\
    Fast-RIR++~\cite{ratnarajah2022fast} & 1.37  & 1.25  & 137   & 13.7  & 1.45  & 1.61  & 369   & 15.2 \\
    \baseline{}~\cite{majumder2022few} & 1.10   & 0.43  & 106   & 8.7  & 1.22  & 0.65  & 164   & 10.5 \\
    \midline
    \shortname~(Ours)  &     \textbf{1.08}  & \textbf{0.42}  & \textbf{90}    & \textbf{8.5}   & \textbf{1.20}  & \textbf{0.61}  & \textbf{153}   & \textbf{10.3} \\
    \bottomline
    \end{tabular}
}
\vspace{-7mm}
\end{table}

\textbf{Results on Matterport3D.} We compare \shortname{} with the aforementioned methods and baselines on the Matterport3D dataset. As shown in Table~\ref{tab:results}, non-learned methods (\ie~Nearest Neighbor and Linear Interpolation) perform poorly on STFT error. We suspect that simply aggregating the RIRs from the neighbors is insufficient to infer the accurate RIRs of the novel viewpoints. The improvements over AnalyticalRIR++ and Fast-RIR++ highlight the rationality of RIR prediction module design. Compared with the state-of-the-art method \baseline{}, our \shortname{} has better performance on all metrics. On the val \textit{unseen} split, \shortname{} outperforms \baseline{} on STFT by 1.64\%, indicating that the RIR  predicted by \shortname{} is better fitted to the ground truth at the acoustic spectrogram level. Moreover, at the waveform level, \shortname{} reduces the error in the estimated energy ratio between direct and reverberant sounds in an RIR on RTE by 6.15\% and DRRE by 6.71\%. 
At the level of perceptual quality, \shortname{} leads to higher generalization performance, surpassing \baseline{} by 1.90\% on MOSE. These results suggest that our \shortname{} learns spatial sounds more in line with human perception and is more satisfying.

\icml{\textbf{Results on Replica.} We also compare \shortname{} with the baselines on Replica. As shown in Table~\ref{tab:replica}, our \shortname{} consistently outperforms all methods. Note that the performance improvements over the baselines are more obvious than on Matterport3D. We attribute the improvement to the \anticipatedmap{} to explicitly reduce the learning difficulty on Replica, as the sampled support observations are more dense and redundant due to the higher spatial resolution. Projecting the features into a map allows the model to encode and understand features from spatial relationships, rather than simply weighting similar features. All these results show the effectiveness of the proposed \shortname{},
which learns a comprehensive map representation for \task{}.}

\subsection{Ablation study}
Considering the time efficiency, we validate the effectiveness of our core proposed modules on the mini Matterport3D dataset which is described in Section~\ref{sec:es}. The results are shown below,

\textbf{Effectiveness of feature mapper.} Projecting multiple encoded features onto the same spatial top-down map provides a clear visual representation of the observations provided. In Table~\ref{tab:ablation}, we perform ablations on the feature mapper module. We construct a variant whose visual observations are purely encoded by the transformer (row 1). We then replace the visual branch with a feature mapper to generate the \originmap{} (row 2). Comparing row 1 and row 2 in Table~\ref{tab:ablation}, the model with feature mapper performs better when exploiting the \originmap{}. We suspect that this is because the \originmap{} makes better use of the available observation information to provide a more integrated representation of the model's predicted RIR.

\textbf{Effectiveness of feature anticipation.} Benefiting from the feature anticipation module, we predict the features of the unseen region based on the features of the given support observations to obtain the whole scene features. In probing the unique contributions of feature anticipation, in Table~\ref{tab:ablation} (row 3), we perform feature anticipating based on the \originmap{}. We find that the performance of the model improved substantially. This can be attributed that when lacking features of unseen regions of the scene, it is harder for the model to learn the acoustic features of the space as a whole, and then predict the RIR between arbitrary locations.

\textbf{Effectiveness of skip connection in the feature anticipation module. } Motivated by ResNet, we aim to reserve the features from the seen area and reduce the learning difficulty by adding the skip connection into the feature anticipation module. To evaluate the effectiveness of the skip connection, we design a variant that removes it from the module. In Table~\ref{tab:skip connection}, this variant performs worse than our ~\shortname{}, MOSE decreasing significantly from 12.9 to 12.3. This shows the necessity of the skip connection, which helps to make U-Net focus on predicting features in unseen regions and reduce the difficulty in feature prediction.

\textbf{Effect of different modal features for constructing the map. } \shortname{} obtains acoustic-related information mainly from visual semantic features. In this subsection, we would like to evaluate whether the map constructed from audio features benefits the few-shot RIR prediction. To this end, we modify the audio branch of \baseline{} to project the audio features into a top-down map. 
In Table~\ref{tab:audio map}, the variant using the visual feature map outperforms the variant using the audio feature map in terms of almost all metrics. 
This could be attributed to the fact that the visual semantic features contain information related to sound propagation (\ the shapes and materials of objects) and are also reasonable to project as a map. 

\subsection{Hyperparameters analysis}
In this section, we proceed with exploring the effect of hyperparameters in our experiments, including map size $m$ and context size $N$.

\begin{table}[!t]
\centering
\caption{RIR prediction results on Replica. All metrics use base $10^{-2}$ and lower is better.}
\label{tab:replica}
\resizebox{0.8\linewidth}{!}{\begin{tabular}{lcccc|cccc}
    \topline
    \multirow{2}[2]{*}{Model} & \multicolumn{4}{c|}{Seen} & \multicolumn
{4}{c}{Unseen} \\
    \multicolumn{1}{l}{} & STFT↓ & \multicolumn{1}{c}{RTE↓} & \multicolumn{1}{c}{DRRE↓} & \multicolumn{1}{c|}{MOSE↓} & \multicolumn{1}{c}{STFT↓} & \multicolumn{1}{c}{RTE↓} & \multicolumn{1}{c}{DRRE↓} & \multicolumn{1}{c}{MOSE↓} \\
    \midline
    Nearest Neighbor & 4.87  & 2.20  & 454   & 21.2  & 4.91  & 0.89  & 459   & 19.6  \\
    Linear Interpolation & 4.63  & 1.37  & 456   & 19.6  & 4.62  & 0.73  & 458   & 18.0  \\
    \baseline{}~\cite{majumder2022few} & 1.70  & 0.51  & 408   & 13.0  & 2.12  & 0.51  & 429   & 13.1  \\ \midline
    \shortname~(Ours) & \textbf{1.36 } & \textbf{0.38 } & \textbf{206 } & \textbf{8.3 } & \textbf{1.78 } & \textbf{0.47 } & \textbf{252 } & \textbf{10.3 } \\
    \bottomline
    \end{tabular}
}
\vspace{-4mm}
\end{table}

\begin{table}[t!]
\centering
\caption{Ablation study on feature mapper and feature anticipation.}
\label{tab:ablation}
\resizebox{0.8\linewidth}{!}{\begin{tabular}{cc|cccc|cccc}
    \topline
    \multicolumn{2}{c|}{Module} & \multicolumn{4}{c|}{Seen}     & \multicolumn{4}{c}{Unseen} \\
    Feature Mapper   & Feature Anticipation & STFT↓  & RTE↓   & DRRE↓  & MOSE↓  & STFT↓  & RTE↓   & DRRE↓  & MOSE↓ \\
    \midline
     \XSolidBrush & \XSolidBrush & 1.85  & 1.55  & 429   & 20.3  & 1.82  & 3.10  & 483   & 20.8 \\
     \checkmark & \XSolidBrush & 1.78  & 1.28  & 339   & 21.2  & 1.69  & 1.54  & 340   & 21.0 \\
      \checkmark & \checkmark & \textbf{1.36}  & \textbf{0.82}  & \textbf{153}   & \textbf{11.7}  & \textbf{1.38}  & \textbf{1.25}  & \textbf{173}   & \textbf{12.3} \\
    \bottomline
    \end{tabular}
}
\vspace{-4mm}
\end{table}

\begin{table}[t!]
  \centering
  \caption{Ablation study on map size of the feature map.}
  \label{tab:map}
  \resizebox{0.8\linewidth}{!}{
    \begin{tabular}{c|cccc|cccc}
    \topline
    \multirow{2}[2]{*}{Map Size} & \multicolumn{4}{c|}{Seen}     & \multicolumn{4}{c}{Unseen} \\
          & \multicolumn{1}{c}{STFT↓} & \multicolumn{1}{c}{RTE↓} & \multicolumn{1}{c}{DRRE↓} & \multicolumn{1}{c|}{MOSE↓} & \multicolumn{1}{c}{STFT↓} & \multicolumn{1}{c}{RTE↓} & \multicolumn{1}{c}{DRRE↓} & \multicolumn{1}{c}{MOSE↓} \\
    \midline
    32    & 1.38  & 1.00  & 221   & 11.9  & 1.41  & 1.28  & 226   & 12.5  \\
    40    & 1.37  & 0.84  & 203   & \textbf{11.7 } & 1.39  & 1.22  & 220   & 12.5  \\
    64    & \textbf{1.36} & \textbf{0.82} & \textbf{153} & 11.8  & \textbf{1.38}  & \textbf{1.25}  & \textbf{173 }  & \textbf{12.3}  \\
    128   &     1.37  & 0.96  & 161   & 11.8  & \textbf{1.38}  & 1.28  & 175   & 12.8  \\
    \bottomline
    \end{tabular}
    }
\vspace{-5mm}
\end{table}

\textbf{Map size.} The feature map stores all the features from observations. We set the default map size to 64 and the resolution to 1. To inspect the effect of the map side, we apply other map sizes to adjust the resolutions, \ie~the larger the map size, the smaller the resolution. The experimental results are shown in Table~\ref{tab:map}. We observe that the best performance occurs when the map size is 64. We suspect that this is because when the map size becomes smaller, the resolution becomes larger, resulting in multiple features being mapped to the same point. This incurs some features being mixed, thus overlooking some potentially useful features. 
If the map is too large then the resolution will be small, it may contain more noisy environmental information which will increase the learning difficulty of the model.

\begin{figure}[t]
  \centering
  \begin{minipage}{0.38\textwidth}
    \centering
	\includegraphics[width=1.0\linewidth]{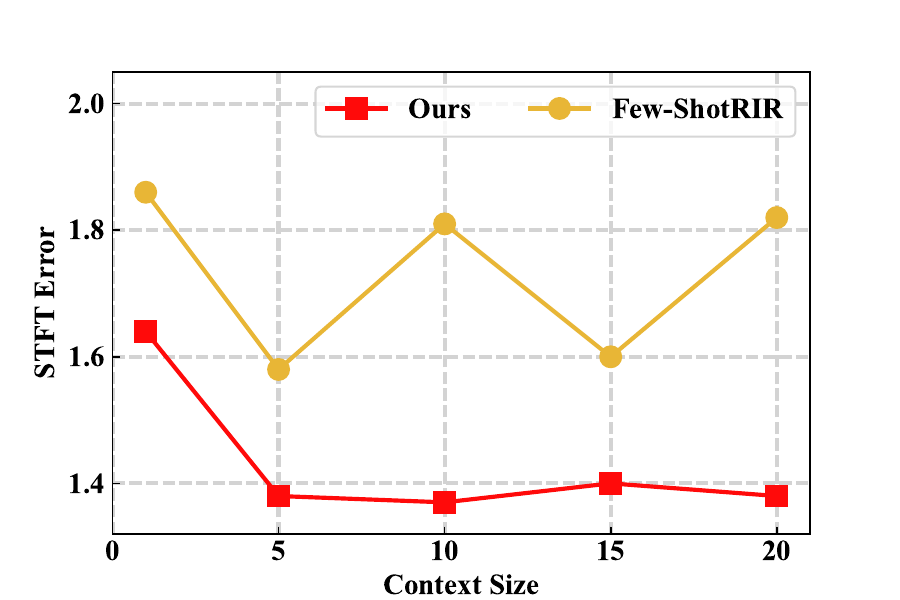}
    \vspace{-6mm}

	\caption{STFT error \vs context size. Comparison of \shortname~(Ours) and \baseline{} on the unseen split.}
    \label{fig:N-figure}
  \end{minipage}
  \hfill
  \begin{minipage}{0.6\textwidth}
    \centering
    \begin{minipage}{\linewidth}
      \centering
      \captionof{table}{Ablation study on skip connection in the feature anticipation module.}
      \label{tab:skip connection}
        \resizebox{1.0\linewidth}{!}{\begin{tabular}{c|cccc|cccc}
            \topline
            \multirow{2}[2]{*}{Skip Connection}  & \multicolumn{4}{c|}{Seen}     & \multicolumn{4}{c}{Unseen} \\
             & STFT↓  & RTE↓   & DRRE↓  & MOSE↓  & STFT↓  & RTE↓   & DRRE↓  & MOSE↓ \\
            \midline
            \XSolidBrush & 1.37  & 1.27  & 158   & 11.9 & 1.39  & 1.75  & 174   & 12.9  \\
            \checkmark  & \textbf{1.36}  & \textbf{0.82}  & \textbf{153}   & \textbf{11.7}  & \textbf{1.38}  & \textbf{1.25}  & \textbf{173}   & \textbf{12.3} \\
            \bottomline
            \end{tabular}
        }
    \end{minipage}
    \begin{minipage}{\linewidth}
      \centering
    \vspace{2mm}
    \captionof{table}{Ablation study on different modal features for constructing the map.}
    \label{tab:audio map}
    \resizebox{1.0\linewidth}{!}{\begin{tabular}{c|cccc|cccc}
        \topline
        \multirow{2}[2]{*}{Feature}  & \multicolumn{4}{c|}{Seen}     & \multicolumn{4}{c}{Unseen} \\
         & STFT↓  & RTE↓   & DRRE↓  & MOSE↓  & STFT↓  & RTE↓   & DRRE↓  & MOSE↓ \\
        \midline
        Audio & 1.39  & 1.78  & \textbf{142}   & 12.4  & 1.45  & 3.04  & \textbf{167}  & 13.7  \\
        Vision  & \textbf{1.36}  & \textbf{0.82}  & 153   & \textbf{11.7}  & \textbf{1.38}  & \textbf{1.25}  & 173   & \textbf{12.3} \\
        \bottomline
        \end{tabular}
    }
    \end{minipage}
  \end{minipage}
\vspace{-2mm}
\end{figure}

\textbf{Context size.} Context size represents the number of support observations given. We decrease $N$ to explore the performance of the model under a more few-shot setting. As shown in Figure~\ref{fig:N-figure}, when $N$ is reduced to 10, there is no very noticeable performance change in \shortname{}. The scene maps can continue to provide scene features for RIR prediction despite less context. The performance of our method experiences significant degradation only when $N$ is set to 1. We argue that this is because the \anticipatedmap{} constructed with a single observation expresses limited information. Notably, our method outperforms \baseline{} under all of the situations regardless of the $N$.

\begin{figure*}[!t]
    \centering
	\includegraphics[width=1.0\linewidth]{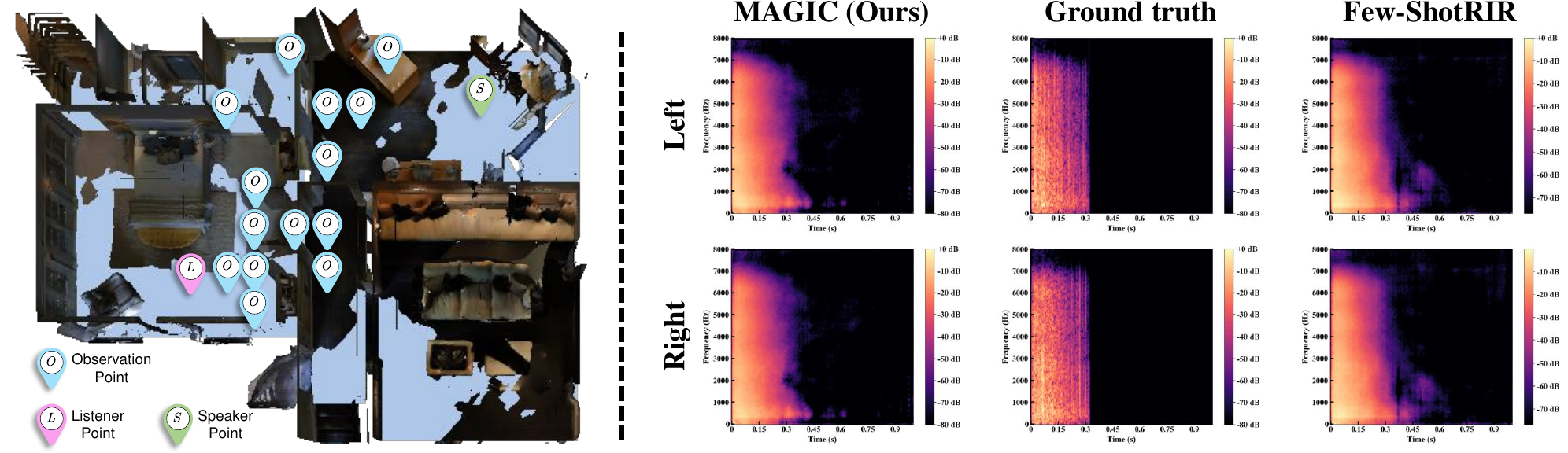}
	\vspace{-6mm}
	\caption{Qualitative RIR prediction. The left half is the top-down view of the scene. The right half shows the predicted RIR of \shortname~(Ours), ground truth, and \baseline{}.}
    \vspace{-4.5mm}
    \label{fig:visualization}
\end{figure*}

\subsection{Qualitative results}

As illustrated in Figure~\ref{fig:visualization}, we visualize a challenging situation when the distance between the coordinates of the query (both speaker and listener) and the provided viewpoints is far. In this situation, \baseline{} can not predict the RIR well as the results it generates have a larger portion of noise at low and medium frequencies, indicating that it is a poor predictor of out-of-range query. In contrast, \shortname{} yields a noticeable improvement compared to the \baseline{}. We attribute this to our proposed \shortname{} constructs a \anticipatedmap{} that actively predicts unseen regions (\eg, features in the region where the speaker and listener are located) and also models the relationship between features. More demos and failure case analyses can be found in supplementary materials.

\section{Conclusion}
In this paper, we propose \shortname, a few-shot acoustic learning paradigm that perceives the environment with a proposed \anticipatedmap{} to capture the inherent correlation between acoustic-related semantic features and sound propagation processes in space. We extract semantic features and then project them to the \originmap{} to help the model build an understanding of given few-shot observations. Considering the feature scarcity brought by few-shot observations, we propose the feature anticipation module to learn the features of unseen regions from seen regions to obtain the expanded \anticipatedmap. We combine the \anticipatedmap{} with echo encoding to an encoder-decoder to predict RIR for arbitrary speaker-listener query pairs. Experimental results demonstrate the efficacy of the proposed method in challenging real-world sounds and environments.

\medskip

{
	\small
	\bibliographystyle{abbrv}
	\bibliography{ref}
}

\def\mytitle{Map-Guided Few-Shot Audio-Visual Acoustics Modeling}

\def\revised{\textcolor{black}}

\appendix
\onecolumn
\begin{center}
	{
		\Large{\textbf{Supplementary Materials for \\  ``Map-Guided Few-Shot Audio-Visual Acoustics Modeling''}}
	}
\end{center}
\vspace{15 pt}

In this supplementary material, we provide more implementation details, more experimental results, more qualitative results, and more discussion of our \shortname. We organize the supplementary as follows:

\begin{itemize}
    \item In Section \ref{sec:Algorithm Procedure}, we provide the training paradigm of \shortname.
    \item In Section \ref{sec:Feature Fusion Details}, we provide explanations on the mechanism of fusing visual features between different levels.
    \item In Section \ref{sec:Depth Mapping Mechanism Details}, we provide more details on the depth mapping mechanism.
    \item In Section \ref{sec:Observation Details}, we provide more details on the observations.
    \item In Section \ref{sec:Model Architecture Details}, we provide more model architecture details of \shortname.
    \item In Section \ref{sec:Dataset Details}, we provide more details of the Matterport3D dataset and Replica dataset during the training phase and inference phase.
    \item In Section \ref{sec:Experimental Details}, we provide more experimental details, including the optimizer hyperparameters and the scheduler hyperparameters.
    \item In Section \ref{sec:Cross-Environment Evaluation}, we provide more results on cross-environment evaluation.
    \item In Section \ref{sec:Qualitative Results}, we provide more qualitative results including demos and failure cases.
    \item In Section \ref{sec:Limitations}, we provide a brief discussion of limitations.
    \item In Section \ref{sec:Broader Impacts}, we provide broader impacts of \shortname.
\end{itemize}

\section{Algorithm procedure}
\label{sec:Algorithm Procedure}

\begin{algorithm}[!ht]
    \small
    \caption{Training paradigm for \shortname}
    \label{alg:training_algorithm}
    \begin{algorithmic}[1]
    \REQUIRE
    The visual encoder $f_V$, the feature mapper $m(\cdot)$, the feature anticipation module $f_{FA}$, the audio encoder $f_A$, the audio fusion module $f_{F}$, the query module $f_{Q}$

    \STATE Initialize the audio modality embedding $E_A$
    
    \FOR{training instances}
        \STATE Initialize the \originmap{} $M_{\mathrm{\originmapshortname{}}} = 0$
        \STATE Initialize the \anticipatedmap{} $M_{\mathrm{\anticipatedmapshortname{}}} = 0$ 
        \STATE Collect observations $\mathbfcal{O}$ from the environment, including the visual observations $\mathbfcal{V}$, the audio observations $\mathbfcal{A}$, and the pose observations $\mathbfcal{P}$
        \STATE Collect the query from the environment, including the query pose $\mathbfcal{Q}$ and the corresponding ground-truth RIR $\mathbfcal{R}$
        \STATE // Map Construction
        \STATE Extract the acoustic-related semantic features $V = f_V(\mathbfcal{V})$
        \STATE Project the acoustic-related semantic features $V$ to the \originmap{} $M_{\mathrm{\originmapshortname{}}} = m(V, \mathbfcal{P})$
        \STATE Generate the \anticipatedmap{} $M_{\mathrm{\anticipatedmapshortname{}}} = f_{FA}(M_{\mathrm{\originmapshortname{}}})$
        \STATE // Audio Branch
        \STATE Extract the audio features $A=f_A(\mathbfcal{A})$
        \STATE Get the audio representation $A' = f_{F}(A, E_A, \mathbfcal{P})
$
        \STATE // Encoder-Decoder
        \STATE Predict the RIR $P = f_{Q}(M_{\mathrm{\originmapshortname{}}}, A', \mathbfcal{Q})$
        \STATE Compute the loss via Equation ~\ref{eq:loss}
        \STATE update the $f_V, f_{FA}, f_A, E_A, f_{F}, f_{Q}$
    \ENDFOR
    \end{algorithmic}
\end{algorithm}

\section{Feature fusion details}
\label{sec:Feature Fusion Details}
The fusion of low-level and high-level features is achieved by the skip connections in U-Net. The skip connections in U-Nets provide a direct pathway for information flow between the corresponding encoder and decoder layers. These connections enable the fusion of features at different levels while maintaining the same dimensional space for both low-level and high-level features.

Specifically, by adding the features obtained from the encoder to the corresponding decoder features, we promote the fusion of information from both low-level and high-level representations. This process enhances feature integration by leveraging both detailed spatial information from low-level features and semantic context from high-level features.
 The skip connections facilitate effective feature fusion without altering the dimensionality of the features and promote comprehensive feature representation, contributing to improved model performance.

Note that the U-Net skip connection strategy, as described, has been utilized in prior research~\cite{chen2022weakly} to obtain more representative features.

\section{Depth mapping mechanism details}
\label{sec:Depth Mapping Mechanism Details}
In equation~\ref{eqa:mapping}, the function $m(\cdot)$ denotes the depth mapping/projection mechanism. Specifically, it maps pixel-wise image features into spatial coordinates based on corresponding depth information. This process involves the following steps:
\begin{enumerate}
    \item Conversion of pixel depth values to vertical coordinates in the spatial bird's eye view.
    \item Conversion of horizontal image coordinates to horizontal coordinates in the spatial bird's eye view.
    \item Alignment and mapping of all visual features onto the same spatial map through rotation and translation.
    \item Aggregation of multiple values for the same point, typically by taking the maximum value, to highlight significant features.
\end{enumerate}

\section{Observation details}
\label{sec:Observation Details}
\textbf{Visual observations.} Visual observations consist of egocentric RGB images and egocentric depth images. The corresponding field of view (FOV) for image acquisition is $90^{\circ}$. The height $h$ and width $w$ of the image is $128$. 

\textbf{Audio observations.} Audio observations are provided in the form of dual-channel echo spectrograms. The form of the sound in the Matterport3D ~\cite{chang2017matterport3d} scene and Replica ~\cite{straub2019replica} provided by SoundSpaces ~\cite{chen2020soundspaces} is a waveform obtained by sampling at a sampling rate of $16 kHz$ and $44.1 kHz$, respectively. We use Fourier transform~\cite{sejdic2009time} to transform the waveform to the sound spectrograms.

\textbf{Pose observations.} Pose observations contain the coordinates of the speaker and listener in 3D scenes, as well as the orientation of the listener. We sample viewpoints at a resolution of 1m, which ensures that the relative values between the coordinates are all integers. There are four types of orientations, including $\theta_i \in \{{0^{\circ}, 90^{\circ}, 180^{\circ}}, 270^{\circ}\}$. For ease of calculation, we use the radian system.

\section{Model architecture details}
\label{sec:Model Architecture Details}
\textbf{Visual encoder.} The visual encoder $f_V$ is a pre-trained U-Net ~\cite{ronneberger2015u} semantic segmentation model. It takes egocentric RGB from the observation set as input and outputs the feature images with constant length and width. The U-Net contains 4 layers of downsampled networks and 4 layers of upsampled features, allowing the image size to be compressed from 128 to 4 and then expanded to 128. the dimensional features are expanded from 3 to 512 and then compressed to 64.

\begin{figure*}[!t]
    \centering
	\includegraphics[width=1.0\linewidth]{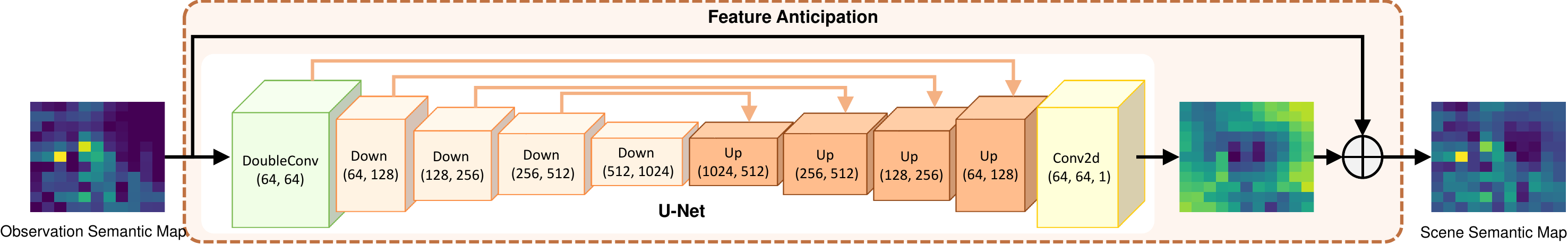}
	\vspace{-2mm}
	\caption{Architecture of feature anticipation module.}
    \vspace{-4mm}
    \label{fig:anticipation}
\end{figure*}

\textbf{Feature anticipation.} The feature anticipation module $f_{FA}$ is combination of U-Net~\cite{ronneberger2015u} model and skip connection ~\cite{he2016deep}. As shown in Figure~\ref{fig:anticipation}, it takes the constructed \originmap{} from egocentric RGB as input and outputs the \anticipatedmap{}. The U-Net contains 4 layers of downsampled networks and 4 layers of upsampled features, allowing the map size to be compressed from 64 to 4 and then expanded to 64. the dimensional features are expanded from 64 to 1024 and then compressed to 64.

\textbf{Audio encoder.} The audio encoder concatenates the echo feature, modality feature, and pose feature, projecting the concatenated embedding with a single linear layer to get audio representation. The echo feature is obtained by encoding the RIR log magnitude spectrogram corresponding to the echoes using the ResNet-18 network~\cite{he2016deep}. The modality feature is a learnable token embedding to distinguish between the visual and audio modalities in the context. The pose feature is the sinusoidal positional embedding of the relative positions of the points. The output is a 512-dimensional feature.

\textbf{Query.} The map embedding from Section~\ref{sec:\anticipatedmap{}} and the audio feature extracted from the audio encoder are concatenated together and fed into the query module. This module takes the query pose as input and predicts the corresponding RIR. The query module is composed of an encoder-decoder structure and a multi-layer network. The encoder-decoder is built with transformer ~\cite{vaswani2017attention}. The encoder uses the self-attention mechanism to learn the relationship between the features and implicitly models the acoustic properties of the 3D scene. The decoder is then used to get the audio features corresponding to the query pose with the help of the cross-attention mechanism. Finally, a multi-layer network is composed of multiple transpose-convolution operations that are used to recover the complete acoustic spectrogram from the audio features. 

\section{Dataset details}
\label{sec:Dataset Details}
\begin{table}[t]
\centering
\caption{Comparision between the different datasets in the experiments.}
\label{tab:dataset}
\resizebox{0.7\linewidth}{!}{\begin{tabular}{c|cc|cc|cc}
    \topline
    \multirow{2}[2]{*}{dataset} & \multicolumn{2}{c|}{train} & \multicolumn{2}{c|}{seen} & \multicolumn{2}{c}{unseen} \\
    \multicolumn{1}{c|}{} & \multicolumn{1}{c}{scene} & \multicolumn{1}{c|}{total} & \multicolumn{1}{c}{scene} & \multicolumn{1}{c|}{total} & \multicolumn{1}{c}{scene} & \multicolumn{1}{c}{total} \\
    \midline
    Matterport3D (complete) & 57    & 11970 & 57    & 798   & 26    & 364 \\
    Matterport3D (mini) & 12    & 600   & 12    & 175   & 26    & 364 \\
    Replica & 9     & 1890  & 9     & 126   & 9     & 126 \\
    \bottomline
    \end{tabular}
}
\end{table}

We validate the effectiveness of our method on three different datasets in Table~\ref{tab:dataset}. We detail the settings of the datasets as below.

\textbf{Matterport3D (complete) dataset.} Following \baseline{}~\cite{majumder2022few}, the scenes are divided into \textit{seen} split and \textit{unseen} split, containing 56 scenes and 27 scenes, respectively. During the training phase, each of the 56 scenes in the \textit{seen} split is utilized 210 times, accumulating a total of 11,970 instances. During the inferencing phase, there are 798 instances in the \textit{seen} split and 364 instances in the \textit{unseen} split to be tested. 

\textbf{Matterport3D (mini) dataset.} We randomly select 12 \textit{seen} scenes from the 56 \textit{seen} scenes for the mini-training setting.  We list all 12 seen scenes as follows: {'YmJkqBEsHnH', 'gTV8FGcVJC9', 'B6ByNegPMKs', 'uNb9QFRL6hY', 'PuKPg4mmafe', 'ZMojNkEp431', '17DRP5sb8fy', 'VFuaQ6m2Qom', '5LpN3gDmAk7', 'V2XKFyX4ASd', 'ac26ZMwG7aT', 'EDJbREhghzL'}. The scenes in \textit{unseen} split remain unchanged.

\textbf{Replica dataset.} We treat 9 sampled ones as seen and the remaining 9 as unseen. During the training phase, each of the 9 scenes in the \textit{seen} split is utilized 210 times, accumulating a total of 1,890 instances. During the inferencing phase, there are 126 instances in both the \textit{seen} split and the \textit{unseen} split to be tested.

\section{Experimental details}
\label{sec:Experimental Details}

\textbf{Optimizer hyperparameters.} For optimization, we utilize Adam optimizer ~\cite{kingma2014adam} with hyperparameters $\beta_1 = 0.9$, $\beta_2 = 0.999$ and $\epsilon= 10^{-5}$. $w_t$ denotes the weight to be updated. $\alpha$ denotes the learning rate. The specific formula is as follows: 
\begin{equation}
m_t = \beta_1 m_{t-1} + (1 - \beta_1) \left(\frac{\partial L}{\partial w_t}\right)
\end{equation}

\begin{equation}
v_t = \beta_2 v_{t-1} + (1 - \beta_2) \left(\frac{\partial L}{\partial w_t}\right)^2
\end{equation}

\begin{equation}
\hat{m}_t = \frac{m_t}{1 - \beta_1^t}
\end{equation}

\begin{equation}
\hat{v}_t = \frac{v_t}{1 - \beta_2^t}
\end{equation}

\begin{equation}
w_{t+1} = w_t - \frac{\alpha}{\sqrt{\hat{v}_t} + \epsilon} \hat{m}_t
\end{equation}

\textbf{Scheduler hyperparameters.} On Matterport3D (complete) dataset, we keep the same settings as \baseline{} and train our model with a fixed learning rate $1.0^{-4}$. On Matterport3D (mini) and Replica datasets, we select a learning rate scheduler with a linear warmup and cooldown which controls the optimizer to make the model converge faster. The learning rate in the learner is $1.0^{-4}$. The percentage of training to perform a linear learning rate warmup is $0.2$. The multiplicative factor applied to the learning rate cooldown slope is $2$.

\section{Cross-environment evaluation}
\label{sec:Cross-Environment Evaluation}
 Cross-environment testing is a comprehensive way to prove model generalizability. We train models on the Replica dataset and test them on the Matterport3D dataset. The results of the experiments on seen and unseen are shown in Table ~\ref{tab:cross}. From the results, our method outperforms \baseline{} on all metrics. The results suggest the generation ability of our method.

 \begin{table}[t!]
\centering
\caption{Generalization results in the cross-environment evaluation. All metrics use base $10^{-2}$ and lower is better.}
\label{tab:cross}
\resizebox{0.8\linewidth}{!}{\begin{tabular}{lcccc|cccc}
    \topline
    \multirow{2}[2]{*}{Model} & \multicolumn{4}{c|}{Seen} & \multicolumn{4}{c}{Unseen} \\
    \multicolumn{1}{l}{} & STFT↓ & \multicolumn{1}{c}{RTE↓} & \multicolumn{1}{c}{DRRE↓} & \multicolumn{1}{c|}{MOSE↓} & \multicolumn{1}{c}{STFT↓} & \multicolumn{1}{c}{RTE↓} & \multicolumn{1}{c}{DRRE↓} & \multicolumn{1}{c}{MOSE↓} \\
    \midline
    \baseline{}~\cite{majumder2022few} & 2.04  & 71.40 & 378.2 & 19.8  & 2.00  & 71.89 & 383.9 & 19.8 \\
    \shortname~(Ours) & \textbf{1.78} & \textbf{60.09} & \textbf{279.3} & \textbf{14.7} & \textbf{1.70} & \textbf{51.75} & \textbf{293.9} & \textbf{16.1} \\
    \bottomline
    \end{tabular}
}
\end{table}

\section{More qualitative results}
\label{sec:Qualitative Results}
We visualize the results of eval and provide the corresponding audio files at \href{https://anonymous.4open.science/r/MAGIC-Preprint/demo.md}{demo}~(please click for checking). Blue pinpoints indicate the provided viewpoints. The blue arrow represents the direction of the provided viewpoints. The green pinpoint indicates the speaker that emits the audio in the query and the pink pinpoint indicates the listener that receives the audio.  In this query, there are provided viewpoints in the vicinity of both the sounding and receiving sources. These viewpoints either coincide with the location of the sounding source or provide a view of the area where the receiving source is located, thus providing a direct visual feature for map construction. Our acoustic-related semantic feature maps explicitly model scene features, and make better use of observational information compared to \baseline{}~\cite{majumder2022few}, resulting in RIR predictions that are closer to the ground truth.

Meanwhile, we visualize a failure result of eval at \href{https://anonymous.4open.science/r/MAGIC-Preprint/failure_case.md}{failure case}~(please click for checking). Our approach relies on constructed acoustic-related semantic feature maps. When the provided visual observations do not have visual features of the query, our prediction results will become poor. It is worth mentioning that the predictive performance of \baseline{}~\cite{majumder2022few} becomes poor in this case as well. For the case where the few-shot observations provided do not cover the entire scene, one possible solution is to train a large scene feature repository to provide additional feature information.

\section{Limitations}
\label{sec:Limitations}
Although our constructed acoustic-related semantic feature maps help to represent the environment comprehensively and provide useful information to improve the accuracy of predicting RIRs, constructing the maps incurs additional memory cost and computational cost. Future work may explore to compress the maps. In addition, while our method has shown promising performance in photo-realistic simulated data, it has not been thoroughly evaluated in the real world. There are various noises in real environments, including sensor noises and driver noises, which tend to interfere with the construction of maps.

\section{Broader impacts}
\label{sec:Broader Impacts}
The ability to predict acoustic-related semantic features and enhance the understanding of the acoustic propagation of a scene has potential applications in the AR domain. Improved acoustic learning could enhance the realism of AR experiences, thus contributing to areas such as gaming, education, and simulation training.

\end{document}